\documentclass[runningheads]{llncs}
\usepackage{graphicx}

\usepackage{cite}
\usepackage{hyperref}
\usepackage[utf8]{inputenc}

\usepackage{framed,multirow}
\usepackage{tikz}
\usetikzlibrary{arrows, shapes, positioning, shadows, trees, calc, decorations.markings}
\usepackage{pifont}
\newcommand{\cmark}{\ding{51}}%
\usepackage{multirow}
\usepackage{booktabs}
\usepackage{amsmath}

\begin{document}
%
\title{SurgicalGPT: End-to-End Language-Vision GPT for Visual Question Answering in Surgery}

\author{Lalithkumar Seenivasan\inst{1, \star} \orcidID{0000-0002-0103-1234} \and
Mobarakol Islam\inst{2, }\thanks{Lalithkumar Seenivasan and Mobarakol Islam are co-first authors.} \orcidID{0000-0002-7162-2822} \and
Gokul Kannan\inst{3} \orcidID{0000-0002-2284-703X} \and
Hongliang Ren\inst{1, 4, 5, }\thanks{Corresponding author.}\orcidID{0000-0002-6488-1551}}

\authorrunning{Seenivasan et al.}

\institute{
Dept. of Biomedical Engineering, National University of Singapore, Singapore. \and
WEISS, University College London, United Kingdom. \and
Dept. of Production Engineering, National Institute of Technology, Tiruchirappalli, India. \and
Dept. of Electronic Engineering, Chinese University of Hong Kong. \and
Shun Hing Institute of Advanced Engineering, Chinese University of Hong Kong.\\
}

%
%
%
%
\maketitle              
\begin{abstract}
Advances in GPT-based large language models (LLMs) are revolutionizing natural language processing, exponentially increasing its use across various domains. Incorporating uni-directional attention, these autoregressive LLMs can generate long and coherent paragraphs. However, for visual question answering (VQA) tasks that require both vision and language processing, models with bi-directional attention or models employing fusion techniques are often employed to capture the context of multiple modalities all at once. As GPT does not natively process vision tokens, to exploit the advancements in GPT models for VQA in robotic surgery, we design an end-to-end trainable Language-Vision GPT (LV-GPT) model that expands the GPT2 model to include vision input (image). The proposed LV-GPT incorporates a feature extractor (vision tokenizer) and vision token embedding (token type and pose). Given the limitations of unidirectional attention in GPT models and their ability to generate coherent long paragraphs, we carefully sequence the word tokens before vision tokens, mimicking the human thought process of understanding the question to infer an answer from an image. Quantitatively, we prove that the LV-GPT model outperforms other state-of-the-art VQA models on two publically available surgical-VQA datasets (based on endoscopic vision challenge robotic scene segmentation 2018 and CholecTriplet2021)  and on our newly annotated dataset (based on the holistic surgical scene dataset). We further annotate all three datasets to include question-type annotations to allow sub-type analysis. Furthermore, we extensively study and present the effects of token sequencing, token type and pose embedding for vision tokens in the LV-GPT model.
\end{abstract}

\section{Introduction} \label{introduction}
The recent evolution of large language models (LLMs) is revolutionizing natural language processing and their use across various sectors (e.g., academia, healthcare, business, and IT) and daily applications are being widely explored. In medical diagnosis, recent works~\cite{wang2023chatcad} have also proposed employing the LLM models to generate condensed reports, interactive explanations, and recommendations based on input text descriptions (predicted disease and report). While the current single-modality (language) LLMs can robustly understand the questions, they still require prior text descriptions to generate responses and are unable to directly infer responses based on the medical image. Although language-only models can greatly benefit the medical domain in language processing, there is a need for robust multi-modality models to process both medical vision and language. In the surgical domain, in addition to the scarcity of surgical experts, their daily schedules are often overloaded with clinical and academic work, making it difficult for them to dedicate time to answer inquiries from students and patients on surgical procedures ~\cite{bates2000error}. Although various computer-assisted solutions~\cite{adams1990computer, rogers1998computer, kneebone2003simulation, sarker2007simulation, hong2021simulation} have been proposed and recorded surgical videos have been made available for students to sharpen their skills and learn from observation, they still heavily rely on surgical experts to answer their surgery-specific questions. In such cases, a robust and reliable surgical visual question answering (VQA) model that can respond to questions by inferring from context-enriched surgical scenes could greatly assist medical students, and significantly reduce the medical expert’s workload~\cite{sharma2021medfusenet}.

In the medical domain, MedfuseNet~\cite{sharma2021medfusenet}, an attention-based model, was proposed for VQA in medical diagnosis. Utilizing the advancements in the transformer models, VisualBert RM~\cite{seenivasan2022surgical}, a modified version of the VisualBert~\cite{li2019visualbert} model was also proposed for VQA in robotic surgery. Compared to most VQA models that require a region proposal network to propose vision patches, the VisualBert RM~\cite{seenivasan2022surgical} performed VQA based on features extracted from the whole image, eliminating the need for a region proposal network. However, they were extracted using a non-trainable fixed feature extractor. While VisualBert~\cite{li2019visualbert} models and LLMs are transformer models, there are fundamentally different. VisualBert~\cite{li2019visualbert} transformers are bidirectional encoder models and are often employed for multi-modality tasks. In contrast, ChatGPT~\footnote{chat.openai.com} (GPT3.5) and BARD (LaMDA~\cite{thoppilan2022lamda}) are language-only uni-directional transformer decoder models employed for language generation. As they are proving to be robust in language generation, exploiting them to process the questions and enabling them to process vision could greatly improve performance in VQA tasks.

In this work, we develop an end-to-end trainable SurgicalGPT model by exploiting a pre-trained LLM and employing a learnable feature extractor to generate vision tokens. In addition to word tokens, vision tokens (embedded with token type and pose embedding) are introduced into the GPT model, resulting in a Language-Vision GPT (LV-GPT) model. Furthermore, we carefully sequence the word and vision tokens to leverage the GPT model’s robust language processing ability to process the question and better infer an answer based on the vision tokens. Through extensive experiments, we show that the SurgicalGPT(LV-GPT) outperforms other state-of-the-art (SOTA) models by $\sim 3-5$ \% on publically available EndoVis18-VQA~\cite{seenivasan2022surgical} and Cholec80-VQA surgical-VQA~\cite{seenivasan2022surgical} datasets.  Additionally, we introduce a novel PSI-AVA-VQA dataset by adding VQA annotations to the publically available holistic surgical scene dataset(PSI-AVA) and observe similar performance improvement. Furthermore, we study and present the effects of token sequencing, where model performance improved by $\sim 2-4$ \% when word tokens are sequenced earlier. Finally, we also study the effects of token type and pose embedding for vision tokens in the LV-GPT model.

\section{Proposed Method} \label{proposed_method}
\subsection{Preliminaries} \label{preliminaries} 

\begin{figure}[!b]
    \centering
    \includegraphics[width=0.83\textwidth]{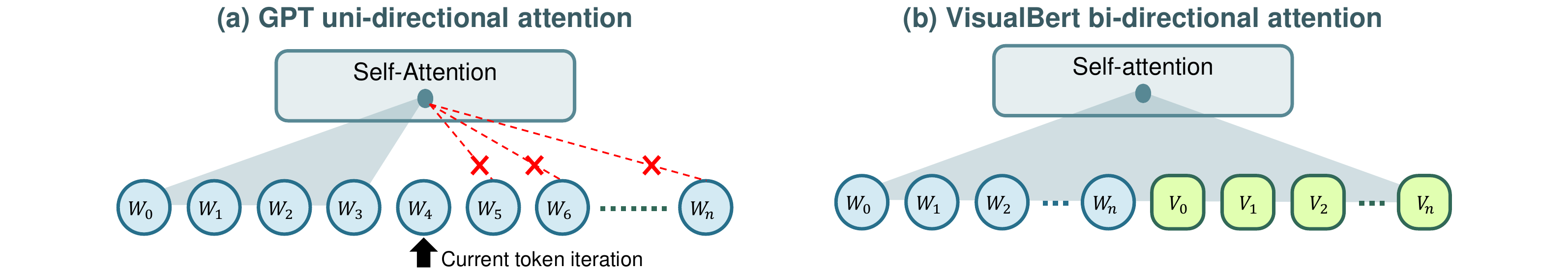}
    \caption{Uni-directional attention in GPT language model vs bi-direction attention in VisualBert multi-modality model.}
    \label{fig:GPTvsVB}
\end{figure}

GPT2~\cite{brown2020language}, a predecessor to GPT3.5 (ChatGPT), is a transformer decoder model that performs next-word prediction. Auto-regressive in nature, its self-attention blocks attend to earlier word tokens to predict the next word token iteratively, allowing the model to generate complex paragraphs~\cite{peng2020soloist}. Although robust in language generation, due to its unidirectional attention~\cite{liu2021gpt}, in a given iteration, the generated token knows all earlier tokens but does not know any subsequent token (Fig.~\ref{fig:GPTvsVB} (a)), restricting the model’s ability to capture the entire context between all tokens. VisualBert~\cite{li2019visualbert}, fundamentally different from GPT models, is a non-auto-regressive transformer encoder model. Its bidirectional self-attention blocks attend in both directions (earlier and subsequent tokens)~\cite{liu2021gpt}, allowing the model to capture the entire context all at once (Fig.~\ref{fig:GPTvsVB} (b)). Due to this, bi-directional attention models are often preferred for multi-modality tasks.

\noindent \textbf{Vision-Language Processing:} Employed mostly for language-only tasks, GPT models do not natively process vision tokens~\cite{guo2022images}. While it supports robust word embedding, it lacks vision tokenizer and vision embedding layers. This limits exploiting its language processing ability for multi-modality tasks. Alternate to GPT, as the VisualBert model is often preferred for multi-modality tasks, it encompasses dedicated embedding layers for both vision and word tokens.

\subsection{LV-GPT: Language-Vision GPT} \label{GPT}

\begin{figure}[!t]
    \centering
    \includegraphics[width=0.93\textwidth]{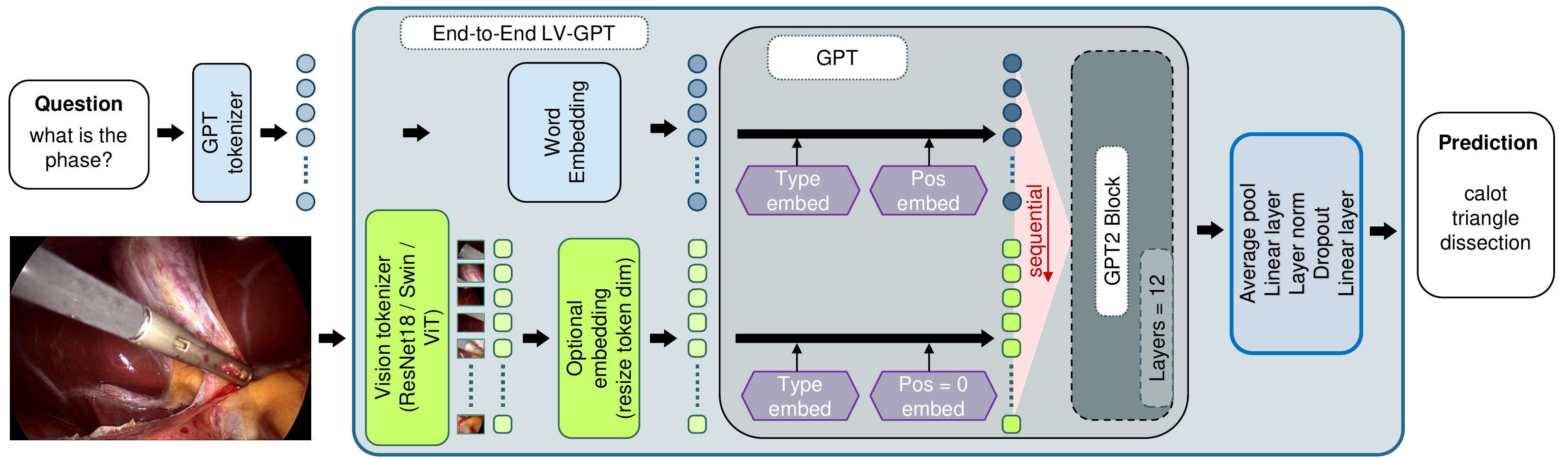}
    \caption{End-to-End LV-GPT for Surgical VQA: The input question and surgical scene are tokenized, embedded, and sequenced to predict the answer. }
    \label{fig:architecture}
\end{figure}

\textbf{Overall Network:}
We design an end-to-end trainable multi-modality (language and vision) LV-GPT model (Fig. \ref{fig:architecture}) for surgical VQA. We integrate a vision tokenizer (feature extractor) module and vision embedding with the GPT model to exploit its language processing ability in performing VQA tasks. 

\noindent \textbf{Language-Vision Processing:} The questions are tokenized using the inherent GPT2 tokenizer. The word tokens are further embedded based on token-id, token type (0) and token position by the inherent GPT2 word embedding layers. To tokenize the input surgical scene (image) into vision tokens, the LV-GPT includes a vision tokenizer (feature extractor): ResNet18 (RN18)~\cite{he2016deep} / Swin~\cite{liu2021swin} / ViT~\cite{dosovitskiy2020image}. Given an image, the tokenizer outputs vision tokens, each holding visual features from an image patch. Additionally, the vision tokens are further embedded based on token type (1) and token position (pos = 0) embeddings. The final embedded word and vision tokens ($w_{e}$ and $v_e$) can be formulated as:

\begin{equation}
\begin{split}
w_{e} &= T_{t=0}(w_{x})+P_{pos}(w_{x})+w_{x}; \quad pos=0,1,2,3,...,n. \\
v_{e} &= T_{t=1}(v_{x})+P_{pos=0}(v_{x})+v_{x}; \quad v_{x} = \left\{\begin{matrix}
v_{t}, &dim(v_t^{i}) = dim(w_{x}^{i}) \\ 
f(v_{t}), &else
\end{matrix}\right.
\end{split}
\label{eq:embedding}
\end{equation}

\noindent where, $T_t()$ is type embedding, $P_{pos}()$ is pose embedding, $w_x$ and $v_x$ are initial word and vision embedding, and $v_t$ are vision tokens. Initial word embeds ($w_x$) are obtained using word embedding based on word token id. Depending on the size ($dim$) of each vision token, they undergo additional linear layer embedding ($f()$) to match the size of the word token. 

\noindent \textbf{Token Sequencing:} LLMs are observed to process long sentences robustly and hold long-term sentence knowledge while generating coherent paragraphs/reports. Considering GPT’s superiority in sequentially processing large sentences and its uni-directional attention, the word tokens are sequenced before the vision tokens. This is also aimed at mimicking human behaviour, where the model understands the question before attending to the image to infer an answer. 

\noindent \textbf{Classification:} Finally, the propagated multi-modality features are then passed through a series of linear layers for answer classification.

\section{Experiment} \label{experiments}

\subsection{Dataset} \label{dataset}

\noindent \textbf{EndoVis18-VQA:}
We employ publically available EndoVis18-VQA~\cite{seenivasan2022surgical} dataset to benchmark the model performance. We use the classification subset that includes classification-based question-and-answer (Q\&A) pairs for $14$ robotic nephrectomy procedure video sequences of the MICCAI Endoscopic Vision Challenge 2018~\cite{allan20202018} dataset. The Q\&A pairs are based on the tissue, actions, and locations of $8$ surgical tools. The dataset includes $11783$ Q\&A pairs based on $2007$ surgical scenes. The answers consist of $18$ classes ($1$ kidney, $13$ tool-tissue interactions, and $4$ tool locations). Additionally, we further annotated the validation set (video sequences 1, 5, and 16) on question types to assist in additional analysis. We followed the EndoVis18-VQA~\cite {seenivasan2022surgical} dataset’s original train/test split.

\noindent \textbf{Cholec80-VQA:}
The classification subset of the Cholec80-VQA~\cite{seenivasan2022surgical} is also employed for model evaluation. It contains Q\&A pairs for $40$ video sequences of the Cholec80 dataset~\cite{twinanda2016endonet}. The subset consists of $43182$ Q\&A pairs on the surgical phase and instrument presence for 21591 frames. The answers include $13$ classes ($2$ instrument states, $4$ on tool count, and $7$ on surgical phase). We additionally annotated the validation set (video sequences: 5, 11, 12, 17, 19, 26, 27 and 31) on the Q\&A pairs types for further model analysis. The VQA~\cite{seenivasan2022surgical} dataset’s original train/test split is followed in this work.

\noindent \textbf{PSI-AVA-VQA:}
We introduce a novel PSI-AVA-VQA dataset that consists of Q\&A pairs for key surgical frames of 8 cases of the holistic surgical scene dataset (PSI-AVA dataset)~\cite{valderrama2022towards}. The questions and answers are generated in sentence form and single-word (class) response form, respectively. They are generated based on the surgical phase, step, and location annotation provided in the  PSI-AVA dataset~\cite{valderrama2022towards}. The PSI-AVA-VQA consists of $10291$ Q\&A pairs and with $35$ answer classes ($4$ locations, $11$ surgical phases, and $21$\footnote[1]{One class shares a common name with a surgical phase class.} surgical steps). The Q\&A pairs are further annotated into 3 types (location, phase, and step). The fold-1 train/test split of parent PSI-AVA~\cite{valderrama2022towards} dataset is followed in this work.

\subsection{Implementation Details}

All variants of our models ~\footnote[2]{Code available: \href{https://github.com/lalithjets/SurgicalGPT}{github.com/lalithjets/SurgicalGPT}} are trained based on cross-entropy loss and optimized using the Adam optimizer. The models were trained for $80$ epoch, with a batch size of 64, except for LV-GPT (ViT) ( batch size = 32 due to GPU limitation). learning rates  lr = $1$x$10^{-5}$, $1$x$10^{-5}$ and $5$x$ 10^{-6}$ are used for EndoVis18-VQA, PSI-AVA-VQA and Cholec80-VQA dataset, respectively. The SOTA VisualBert~\cite{li2019visualbert} and VisualBert RM~\cite{seenivasan2022surgical} models were implemented using their official code repositories. The Block~\cite{ben2019block}, MUTAN~\cite{ben2017mutan}, MFB~\cite{yu2017multi} and MFH~\cite{yu2018beyond} were implemented using the official codes of Block~\cite{ben2019block}.

\section{Results}

\begin{table}[!t]
    \centering
    \caption{Quantitaive comparison of our LV-GPT (Swin), LV-GPT (RN18), and (LV-GPT (ViT) against state-of-the-art models.}
    \scalebox{0.80}{
    \begin{tabular}{c|ccc|ccc|ccc}
    \toprule
    \multirow{2}{*}{\textbf{MODELS}} & \multicolumn{3}{c|}{\textbf{EndoVis18-VQA~\cite{seenivasan2022surgical}}}                                                                 & \multicolumn{3}{c|}{\textbf{Cholec80-VQA~\cite{seenivasan2022surgical}}}                                                                 & \multicolumn{3}{c}{\textbf{PSI-AVA-VQA}}                                                             \\ \cline{2-10} 
                                     & \multicolumn{1}{c|}{\textbf{Acc}}    & \multicolumn{1}{c|}{\textbf{Recall}} & \multicolumn{1}{c|}{\textbf{FScore}} & \multicolumn{1}{c|}{\textbf{Acc}}    & \multicolumn{1}{c|}{\textbf{Recall}} & \multicolumn{1}{c|}{\textbf{FScore}} & \multicolumn{1}{c|}{\textbf{Acc}}    & \multicolumn{1}{c|}{\textbf{Recall}} & \multicolumn{1}{c}{\textbf{FScore}} \\
    \midrule
    \textbf{VisualBert~\cite{li2019visualbert}}              & \multicolumn{1}{c|}{0.6143} & \multicolumn{1}{c|}{0.4282}  & 0.3745                      & \multicolumn{1}{c|}{0.9007} & \multicolumn{1}{c|}{0.6294}  & 0.6300                      & \multicolumn{1}{c|}{0.5853} & \multicolumn{1}{c|}{0.3307}  & 0.3161                     \\ 
    \textbf{VisualBert RM~\cite{seenivasan2022surgical}}       & \multicolumn{1}{c|}{0.6190}       & \multicolumn{1}{c|}{0.4079}        &   0.3583                          & \multicolumn{1}{c|}{0.9001}       & \multicolumn{1}{c|}{0.6573}        &  0.6585                           & \multicolumn{1}{c|}{0.6016}       & \multicolumn{1}{c|}{0.3242}        & 0.3165                          \\ 
    \textbf{Block~\cite{ben2019block}}                   & \multicolumn{1}{c|}{0.6088}       & \multicolumn{1}{c|}{0.4884}        &  0.4470                           & \multicolumn{1}{c|}{0.8948}       & \multicolumn{1}{c|}{0.6600}        & 0.6413                            & \multicolumn{1}{c|}{0.5990}       & \multicolumn{1}{c|}{\textbf{0.5136}}        &  \textbf{0.4933}                          \\ 
    \textbf{Mutan~\cite{ben2017mutan}}                     & \multicolumn{1}{c|}{0.6303}       & \multicolumn{1}{c|}{\textbf{0.4969}}        & \underline{0.4565}                            & \multicolumn{1}{c|}{0.8699}       & \multicolumn{1}{c|}{0.6332}        & 0.6106                            & \multicolumn{1}{c|}{0.4971}       & \multicolumn{1}{c|}{0.3912}        & 0.3322                           \\ 
    \textbf{MFB~\cite{yu2017multi}}                     & \multicolumn{1}{c|}{0.5238}       & \multicolumn{1}{c|}{0.4205}        & 0.3622                            & \multicolumn{1}{c|}{0.8410}       & \multicolumn{1}{c|}{0.5303}        & 0.4588                            & \multicolumn{1}{c|}{0.5712}       & \multicolumn{1}{c|}{\underline{0.4379}}        & \underline{0.4066}                           \\
    \textbf{MFH~\cite{yu2018beyond}}                     & \multicolumn{1}{c|}{0.5876}       & \multicolumn{1}{c|}{0.4835}        & 0.4224                            & \multicolumn{1}{c|}{0.8751}       & \multicolumn{1}{c|}{0.5903}        & 0.5567                            & \multicolumn{1}{c|}{0.4777}       & \multicolumn{1}{c|}{0.2995}        & 0.2213                           \\
    \midrule
    \textbf{LV-GPT (Swin)}             & \multicolumn{1}{c|}{0.6613} & \multicolumn{1}{c|}{0.4460}  & 0.4537                      & \multicolumn{1}{c|}{\textbf{0.9429}} & \multicolumn{1}{c|}{\textbf{0.7339}}  & \textbf{0.7439}                      & \multicolumn{1}{c|}{\underline{0.6033}}       & \multicolumn{1}{c|}{0.4137}        & 0.3767                           \\ 
    \textbf{LV-GPT (RN18)}              & \multicolumn{1}{c|}{\textbf{0.6811}} & \multicolumn{1}{c|}{0.4649}  & \textbf{0.4649}                      & \multicolumn{1}{c|}{0.8746} & \multicolumn{1}{c|}{0.5747}  & 0.5794                      & \multicolumn{1}{c|}{0.5933}       & \multicolumn{1}{c|}{0.3183}        &  0.3168                          \\ 
    \textbf{LV-GPT (ViT)}              & \multicolumn{1}{c|}{\underline{0.6659}} & \multicolumn{1}{c|}{\underline{0.4920}}  & 0.4336                      & \multicolumn{1}{c|}{\underline{0.9232}} & \multicolumn{1}{c|}{\underline{0.6833}}  & \underline{0.6963}                      & \multicolumn{1}{c|}{\textbf{0.6549}}       & \multicolumn{1}{c|}{0.4132}        & 0.3971                           \\ 
    \bottomrule
    \end{tabular}
    }
    \label{table:sota}
\end{table}

\begin{figure}[!t]
    \centering
    \includegraphics[width=1.0\textwidth]{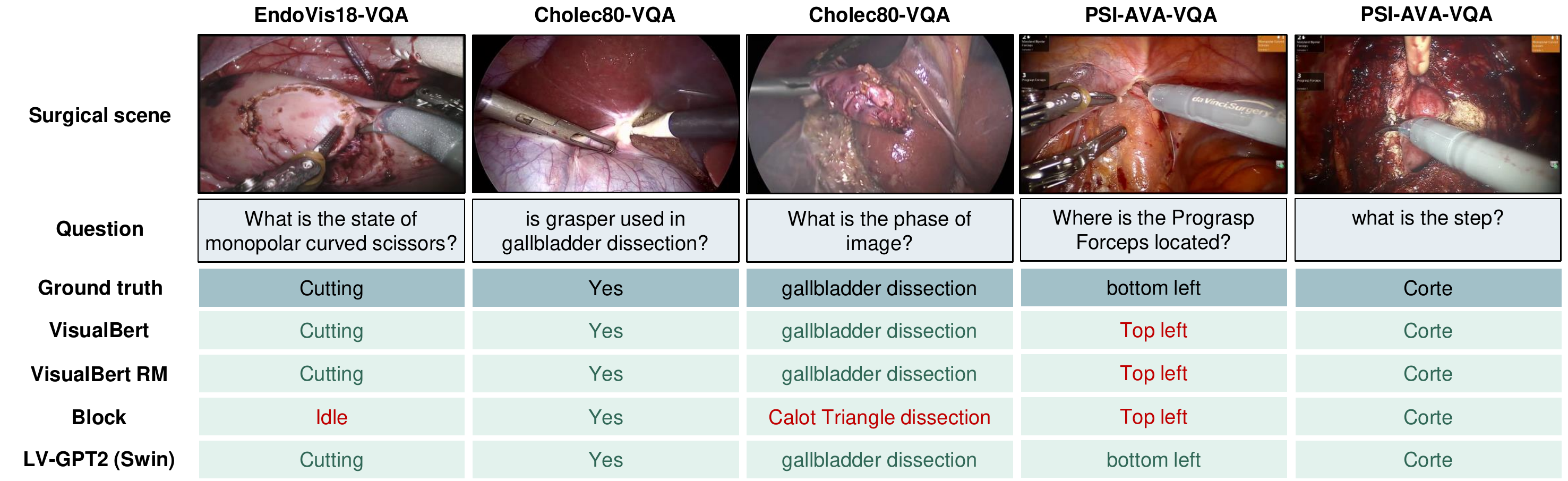}
    \caption{Qualitative analysis: Comparison of answers predicted by VisualBERT~\cite{li2019visualbert}, VisualBert RM~\cite{seenivasan2022surgical}, Block~\cite{ben2019block}, and our LV-GPT (Swin) models against the ground truth based on input surgical scene and question.}
    \label{fig:qualitative_analysis}
\end{figure}

All our proposed LV-GPT model variants are quantitatively benchmarked (Table~\ref{table:sota}) against other attention-based/bi-directional encoder-based SOTA models on EndoVis18-VQA, Cholec80-VQA and PSI-AVA-VQA datasets based on the accuracy (Acc), recall, and Fscore. In most cases, all our variants, LV-GPT (Swin), LV-GPT (RN18) and LV-GPT (ViT), are observed to significantly outperform SOTA models on all three datasets in terms of Acc. Specifically, the LV-GPT (Swin) variant (balanced performance across all datasets) is observed to outperform all SOTA models on all datasets and significantly improve the performance ($\sim 3-5$\% improvement) on EndoVis18-VQA and Cholec80-VQA dataset. Additionally, it should be noted our model variants can be trained end-to-end, whereas, most of the SOTA models requires a region proposal network to process input image into vision tokens. Fig.~\ref{fig:qualitative_analysis} shows the qualitative performance of LV-GPT (Swin) against SOTA models on three datasets. A Comparison of our LV-GPT model performance on the EndoVis18-VQA dataset with default test queries vs rephrased test queries is presented in supplementary materials that highlight the model's robustness in language reasoning.

\subsubsection{Early Vision vs Early Word:}
The performance of LV-GPT based on word and vision token sequencing (Table~\ref{table:tokensequencing}) is also studied. While all three variants of the LV-GPT models processing vision tokens earlier are observed to perform on par with SOTA models reported in Table~\ref{table:sota}, in most cases, their performances on both datasets further improved by $\sim 2-4$\% when word tokens are processed earlier. This improvement could be attributed to LLM’s ability to hold sentence (question) context before processing the vision tokens to infer an answer. This behaviour, in our view, mimics the human thought process, where we first understand the question before searching for an answer from an image.

\begin{table}[!t]
    \centering
    \caption{Comparison of LV-GPT model performance when vision tokens are sequenced earlier vs when word tokens are sequenced earlier.}
    \scalebox{0.80}{
    \begin{tabular}{c|c|ccc|ccc}
        \toprule
        \multirow{2}{*}{\textbf{Token sequencing}} & \multirow{2}{*}{\textbf{Model}} & \multicolumn{3}{c|}{\textbf{EndoVis18-VQA}}                                                  & \multicolumn{3}{c}{\textbf{PSI-AVA-VQA}}                                                      \\ \cline{3-8} 
                                                   &                                 & \multicolumn{1}{c|}{\textbf{Acc}}    & \multicolumn{1}{c|}{\textbf{Recall}} & \textbf{FScore} & \multicolumn{1}{c|}{\textbf{Acc}}    & \multicolumn{1}{c|}{\textbf{Recall}} & \textbf{FScore} \\
        \midrule
        \multirow{3}{*}{\textbf{Early vision}}     & \textbf{LV-GPT (RN18)}         & \multicolumn{1}{c|}{0.6338}          & \multicolumn{1}{c|}{0.3600}          & 0.3510          & \multicolumn{1}{c|}{0.5542}          & \multicolumn{1}{c|}{0.2879}          & 0.2886          \\ 
                                                   & \textbf{LV-GPT (Swin)}         & \multicolumn{1}{c|}{0.6208}                & \multicolumn{1}{c|}{0.4059}                & 0.3441                & \multicolumn{1}{c|}{\textbf{0.6068}} & \multicolumn{1}{c|}{\textbf{0.4195}} & \textbf{0.3813} \\ 
                                                   & \textbf{LV-GPT (ViT)}          & \multicolumn{1}{c|}{0.6493}                & \multicolumn{1}{c|}{0.4362}                & 0.3701               & \multicolumn{1}{c|}{0.6023}          & \multicolumn{1}{c|}{0.2802}          & 0.2628          \\ 
        \midrule
        \multirow{3}{*}{\textbf{Early word}}       & \textbf{LV-GPT (RN18)}         & \multicolumn{1}{c|}{\textbf{0.6811}} & \multicolumn{1}{c|}{\textbf{0.4649}} & \textbf{0.4649} & \multicolumn{1}{c|}{\textbf{0.5933}} & \multicolumn{1}{c|}{\textbf{0.3183}} & \textbf{0.3168} \\ 
                                                   & \textbf{LV-GPT (Swin)}         & \multicolumn{1}{c|}{\textbf{0.6613}} & \multicolumn{1}{c|}{\textbf{0.4460}} & \textbf{0.4537} & \multicolumn{1}{c|}{0.6033}          & \multicolumn{1}{c|}{0.4137}          & 0.3767          \\ 
                                                   & \textbf{LV-GPT (ViT)}          & \multicolumn{1}{c|}{\textbf{0.6659}} & \multicolumn{1}{c|}{\textbf{0.4920}} & \textbf{0.4336} & \multicolumn{1}{c|}{\textbf{0.6549}} & \multicolumn{1}{c|}{\textbf{0.4132}} & \textbf{0.3971} \\ 
        \bottomrule
    \end{tabular}}
    \label{table:tokensequencing}
\end{table}

\begin{table}[!b]
    \centering
    \caption{Comparison of model performances on EndoVis18-VQA, Cholec80-VQA and PSI-AVA-VQA datasets when vision tokens are embedded with zero-positional embedding vs actual pose embedding.}
    \scalebox{0.80}{
    \begin{tabular}{c|c|ccc|ccc}
        \toprule
        \multirow{2}{*}{\textbf{Dataset}} & \textbf{Model}          & \multicolumn{3}{c|}{\textbf{Zero Pose Embedding}}                            & \multicolumn{3}{c}{\textbf{Actual Pose Embedding}}                  \\ \cline{2-8} 
                                          & \textbf{Best LV-GPT}   & \multicolumn{1}{c|}{\textbf{Acc}}             & \multicolumn{1}{c|}{\textbf{Recall}}          & \textbf{FScore}          & \multicolumn{1}{c|}{\textbf{Acc}}    & \multicolumn{1}{c|}{\textbf{Recall}}          & \textbf{FScore}          \\
        \midrule
        \textbf{EndoVis18-VQA}           & \textbf{LV-GPT (RN18)} & \multicolumn{1}{c|}{\textbf{0.6811}} & \multicolumn{1}{c|}{0.4649}          & 0.4649          & \multicolumn{1}{c|}{\textbf{0.6811}} & \multicolumn{1}{c|}{\textbf{0.4720}} & \textbf{0.4681} \\ 
        \textbf{Cholec80-VQA}             & \textbf{LV-GPT (Swin)} & \multicolumn{1}{c|}{\textbf{0.9429}} & \multicolumn{1}{c|}{\textbf{0.7339}} & \textbf{0.7439} & \multicolumn{1}{c|}{0.9414} & \multicolumn{1}{c|}{0.7251}          & 0.7360          \\ 
        \textbf{PSI-AVA-VQA}              & \textbf{LV-GPT (ViT)}  & \multicolumn{1}{c|}{\textbf{0.6549}} & \multicolumn{1}{c|}{\textbf{0.4132}} & \textbf{0.3971} & \multicolumn{1}{c|}{0.5905} & \multicolumn{1}{c|}{0.3742}          & 0.3463          \\ 
        \bottomrule
    \end{tabular}}
    \label{table:vision_pos_embed}
\end{table}

\subsubsection{Pose Embedding for Vision tokens:}

\begin{figure}[!b]
    \centering
    \includegraphics[width=1.0\textwidth]{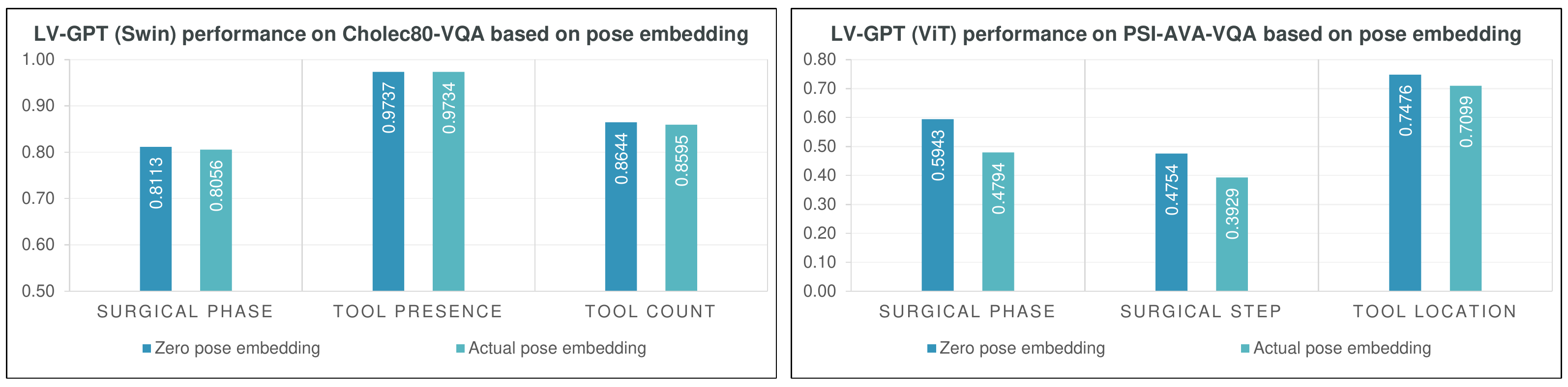}
    \caption{Sub-type performance analysis of LV-GPT model variants on Cholec80-VQA and PSI-AVA-VQA embedded with zero-pose embedding vs actual-pose embedding.}
    \label{fig:subtype}
\end{figure}

The influence of positional embedding of the vision tokens (representing a patch region) in all the LV-GPT variants is studied by either embedded with position information (pos = 1, 2, 3, .., n.) or zero-position (pos = 0). Table~\ref{table:vision_pos_embed} shows the difference in the performance of the best-performing LV-GPT variant in each dataset, with its vision tokens embedded with actual-position or zero-position. While we expected the positional embedding to improve the performance (dataset Q\&A pairs related to tool location), from the results, we observe that embedding vision tokens with zero-position embedding results in better performance. In-depth analysis shows that our CNN-based LV-GPT (RN18) model improved with positional embedding (Table~\ref{table:vision_pos_embed} and Table~\ref{table:ablation}). In the transformer-based LV-GPT (Swin) / LV-GPT (ViT) models, positional embedding is already incorporated at the vision tokenizer (VIT/Swin) layer, and adding positional embedding at the GPT level results in double Position embedding. Thus, “zero-position” can be interpreted as “LV-GPT only requires one layer of positional embedding”. A sub-type analysis (Fig.~\ref{fig:subtype}) is also performed on the model performance to analyze the effect of positional embedding of the vision tokens. The model in which the vision tokens were embedded with zero-position (at the GPT level), performed marginally better/similar on all sub-types in the Cholec80-VQA dataset. However, its performance improvement was significant in the PSI-AVA-VQA dataset sub-types, including the 'tool location' sub-types that contain questions on tool location. 

\subsubsection{Ablation Study on Vision Token Embedding:}
An ablation study on the vision token embedding in the LV-GPT model on the EndoVis18-VQA dataset is also shown in Table~\ref{table:ablation}. VB-VE refers to vision token embedding using VisualBert vision embedding. The C-VE refers to custom embedding, where, in LV-GPT (RN18), the vision token undergoes additional linear layer embedding to match the word-token dimension, and in other variants, vision tokens from the Swin/VIT are directly used. The subsequent VT-TY + VT-PE and VT-TY + VT-ZPE refers to the additional vision token type (TY) and actual-position (PE)/zero-position (ZPE) embedding. We observe that employing C-VE with VT-TY + VT-ZPE results in better performance.

\begin{table}[!t]
    \centering
    \caption{Ablation study on vision token (VT) embedding.}
    \scalebox{0.8}{
    \begin{tabular}{c|c|c|c|clc|ccc}
        \toprule
         \multirow{2}{*}{\textbf{VB-VE}} & \multirow{2}{*}{\textbf{C-VE}} & \multirow{2}{*}{\textbf{\begin{tabular}[c]{@{}c@{}}VT-TY +\\ VT-PE\end{tabular}}} & \multirow{2}{*}{\textbf{\begin{tabular}[c]{@{}c@{}}VT-TY + \\ VT-ZPE\end{tabular}}} & \multicolumn{3}{c|}{\textbf{LV-GPT (RN18)}}                                                   & \multicolumn{3}{c}{\textbf{LV-GPT (ViT)}}                                                    \\ \cline{5-10} 
                                        &                                &                                                                                   &                                                                                     & \multicolumn{1}{c|}{\textbf{Acc}}    & \multicolumn{1}{l|}{Recall}          & \textbf{FScore} & \multicolumn{1}{c|}{\textbf{Acc}}    & \multicolumn{1}{l|}{Recall}          & \textbf{FScore} \\
        \midrule
        \cmark                                &                                &                                                                                   &                                                                                     & \multicolumn{1}{c|}{0.6287}          & \multicolumn{1}{l|}{0.4061}          & 0.4063          & \multicolumn{1}{c|}{0.6147}          & \multicolumn{1}{c|}{0.4199}          & 0.3679          \\ \hline
                                         &  \cmark                              &                                                                                   &                                                                                     & \multicolumn{1}{c|}{0.6728}          & \multicolumn{1}{l|}{0.4366}          & 0.4455          & \multicolumn{1}{c|}{0.6504}          & \multicolumn{1}{c|}{0.4792}          & 0.4323          \\ \hline
                                         & \cmark                               & \cmark                                                                                  &                                                                                     & \multicolumn{1}{c|}{\textbf{0.6811}} & \multicolumn{1}{l|}{\textbf{0.4720}} & \textbf{0.4681} & \multicolumn{1}{c|}{0.6259}          & \multicolumn{1}{c|}{0.4306}          & 0.3805          \\ \hline
                                         &  \cmark                              &                                                                                   &  \cmark                                                                                   & \multicolumn{1}{c|}{\textbf{0.6811}} & \multicolumn{1}{c|}{0.4649}          & 0.4649          & \multicolumn{1}{c|}{\textbf{0.6659}} & \multicolumn{1}{c|}{\textbf{0.4920}} & \textbf{0.4336} \\ 
        \bottomrule
    \end{tabular}}
    \label{table:ablation}
\end{table}

\section{Conclusion}

We design an end-to-end trainable SurgicalGPT, a multi-modality Language-Vision GPT model, for VQA tasks in robotic surgery. In addition to GPT’s inherent word embeddings, it incorporates a vision tokenizer (trainable feature extractor) and vision token embedding (type and pose) to perform multi-modality tasks. Furthermore, by carefully sequencing the word tokens earlier to vision tokens, we exploit GPT’s robust language processing ability, allowing the LV-GPT to significantly perform better VQA. Through extensive quantitative analysis, we show that the LV-GPT outperforms other SOTA models on three surgical-VQA datasets and sequencing word tokens early to vision tokens significantly improves the model performance. Furthermore, we introduce a novel surgical-VQA dataset by adding VQA annotations to the publically available holistic surgical scene dataset. While multi-modality models that process vision and language are often referred to as “vision-language” models, we specifically name our model “language-vision GPT” to highlight the importance of the token sequencing order in GPT models. Integrating vision tokens into GPT also opens up future possibilities of generating reports directly from medical images/videos.

\section{Acknowledgement}
This work was supported by Hong Kong Research Grants Council (RGC) Collaborative Research Fund (CRF C4026-21GF and CRF C4063-18G) and Shun Hing Institute of Advanced Engineering (BME-p1-21/8115064) at the Chinese University of Hong Kong. M. Islam was funded by EPSRC grant [EP/W00805X/1].

\bibliographystyle{splncs04}
\bibliography{paper1492}

\end{document}


%

\title{SurgicalGPT: End-to-End Language-Vision GPT for Visual Question Answering in Surgery}
\titlerunning{Supplementary Material}
%

\newif\ifreview
\reviewtrue

\ifreview
\author{Supplementary Material}
\institute{--}
\fi


%
\maketitle              

\appendix

\begin{table}[]
    \centering
    \caption{Comparison of LV-GPT model performance on EndoVis18-VQA dataset with default test queries vs rephrased test queries to highlight the model's robustness in language reasoning.}
    \begin{tabular}{c|ccc|ccc}
    \toprule
    \multirow{2}{*}{\textbf{Model}} & \multicolumn{3}{c|}{\textbf{Default Test Queries}}                                        & \multicolumn{3}{c}{\textbf{Rephrased Test Queries}}                                    \\ \cline{2-7} 
                                    & \multicolumn{1}{c|}{\textbf{Acc}}    & \multicolumn{1}{c|}{\textbf{Recall}} & \textbf{Fscore} & \multicolumn{1}{c|}{\textbf{Acc}} & \multicolumn{1}{c|}{\textbf{Recall}} & \textbf{Fscore} \\ 
    \midrule
    \textbf{LV-GPT (RN18)}          & \multicolumn{1}{c|}{\textbf{0.6811}} & \multicolumn{1}{c|}{\textbf{0.4649}} & \textbf{0.4649} & \multicolumn{1}{c|}{0.5980}       & \multicolumn{1}{c|}{0.3853}          & 0.3861          \\ \hline
    \textbf{LV-GPT (ViT)}           & \multicolumn{1}{c|}{\textbf{0.6659}} & \multicolumn{1}{c|}{\textbf{0.4920}} & \textbf{0.4336} & \multicolumn{1}{c|}{0.6305}       & \multicolumn{1}{c|}{0.4662}          & 0.4127          \\ \hline
    \textbf{LV-GPT (Swin)}          & \multicolumn{1}{c|}{\textbf{0.6613}} & \multicolumn{1}{c|}{\textbf{0.4460}} & \textbf{0.4537} & \multicolumn{1}{c|}{0.6056}       & \multicolumn{1}{c|}{0.4053}          & 0.4133          \\
    \bottomrule
    \end{tabular}
    \label{Supp_table1}
\end{table}

\bibliographystyle{splncs04}